\documentclass[ASNA,twocolumn]{USG}

\usepackage{amsmath}
\usepackage{amssymb}
\usepackage{booktabs}
\usepackage{xcolor}
\usepackage{tikz}
\usetikzlibrary{arrows.meta,positioning,fit,shapes.geometric}
\usepackage{listings}
\usepackage{siunitx}
\usepackage{multirow}
\usepackage{pifont}
\usepackage{diagbox}

\graphicspath{{./pic/}{./images/}}
\articletype{}
\journal{}
\articledoi{} 
\historydates{} 
\begin{document}
\title{Lang2Morph: Language-Driven Morphological Design of Robotic Hands}
\author[1]{Yanyuan Qiao}
\author[1]{Kieran Gilday}
\author[2]{Yutong Xie}
\author[1]{Josie Hughes}
\authormark{Qiao et al.}
\titlemark{Lang2Morph: Language-Driven Morphological Design of Robotic Hands}
\address[1]{CREATE Lab, Swiss Federal Institute of Technology Lausanne (EPFL), Lausanne, Switzerland}
\address[2]{Computer Vision Department, Mohamed bin Zayed University of Artificial Intelligence (MBZUAI), Abu Dhabi, UAE}
\corres{Yanyuan Qiao (\email{yanyuan.qiao@epfl.ch})}

\abstract[ABSTRACT]{Designing task-specific robotic hands requires translating diverse task requirements into morphologies that are both functional and manufacturable. Existing parametric and optimization-based approaches provide systematic design tools, but often rely on expert-defined objectives, manual tuning, or simulation-based search. Large language models (LLMs), with broad semantic knowledge and reasoning capabilities, offer a promising way to interpret task requirements directly from natural language. We present Lang2Morph, an LLM-driven framework that maps natural-language task descriptions to fabricable robotic hand morphologies within the Open Parametric Hand (OPH) design space. Lang2Morph combines task reasoning, structured morphology generation, geometric constraints, and feedback-guided selection to produce task-conditioned hand designs. Morphology-level analyses demonstrate high design feasibility and task-dependent structural variation, while real-world experiments with three fabricated hands reveal distinct functional characteristics across manipulation tasks. These results highlight the potential of LLMs as practical interfaces for task-driven robotic hand design.}
\maketitle

\section{INTRODUCTION}

Designing the morphology and structure of a robotic hand for a specific application is a fundamental yet long-standing challenge in robotics.  
Unlike general-purpose manipulators, such as anthropomorphic robotic hands, task-specific robotic hands must balance dexterity and generality against optimization for a specific task, while remaining manufacturable~\cite{odhner2013ihy,liu2008dlrhit}.
Designing dexterous yet task-specific hands requires understanding both the manipulation task and fabrication constraints. For example, increasing finger count or joint complexity may improve manipulation capability, but also complicates fabrication and control. As a result, task-specific hand design has typically relied on expert-driven heuristics and iterative prototyping, which are time-consuming and difficult to scale.

The development of parametric hand designs can provide a structured space for systematic exploration of task-specific morphologies.
Open-source hardware projects such as the Yale OpenHand Project~\cite{ma2013openhand,ma2017openhand2} have released reproducible underactuated hand designs, lowering the barrier for building functional prototypes. Parametric frameworks like the Open Parametric Hand (OPH)~\cite{gilday2025oph} further expose a structured design space with tunable parameters, enabling systematic exploration of hand morphologies.
However, both approaches still depend on human designers to interpret tasks, identify likely grasp types, and manually adjust parameters. Without automated mapping from task requirements to design instantiations, scaling to diverse, task-specific morphologies remains difficult.

In parallel, the robotics community has explored automated robot morphology optimization and generation through evolutionary optimization, grammar-based synthesis, and differentiable pipelines~\cite{cheney2013unshackling,robogrammar,task2morph}.  
These works demonstrate that morphology can be generated algorithmically, but they often rely on computationally expensive optimization, physics-based simulation, and carefully designed task-specific objectives or fitness functions. In task-specific hand design, functional requirements such as fingertip precision, lateral pinching, or stabilizing support must therefore be translated into explicit optimization criteria. This motivates exploring whether such requirements can also be interpreted more directly from natural-language descriptions.

Recently, large language models (LLMs) have been increasingly explored in robotics, with applications spanning planning, navigation, and manipulation~\cite{ahn2022saycan,Zhou2023NavGPTER,driess2023palme}. Beyond high-level decision making, LLMs have also begun to influence robot embodiment and morphology design by interpreting task descriptions and guiding the generation of physical structures~\cite{text2robot,fang2025robomore,lin2025robotsmith}. Their broad background knowledge about objects, human--object interactions, and engineering concepts makes them particularly promising for connecting task requirements with physical design choices. Existing studies have mainly focused on locomotion-oriented robots, modular or soft bodies, and task-specific robotic tools. In contrast, LLM-driven robotic hand morphology design remains largely unexplored, despite requiring similarly rich reasoning about task semantics, object properties, and grasp requirements.

\begin{figure}[t]
   \centering
   \includegraphics[width=1\linewidth]{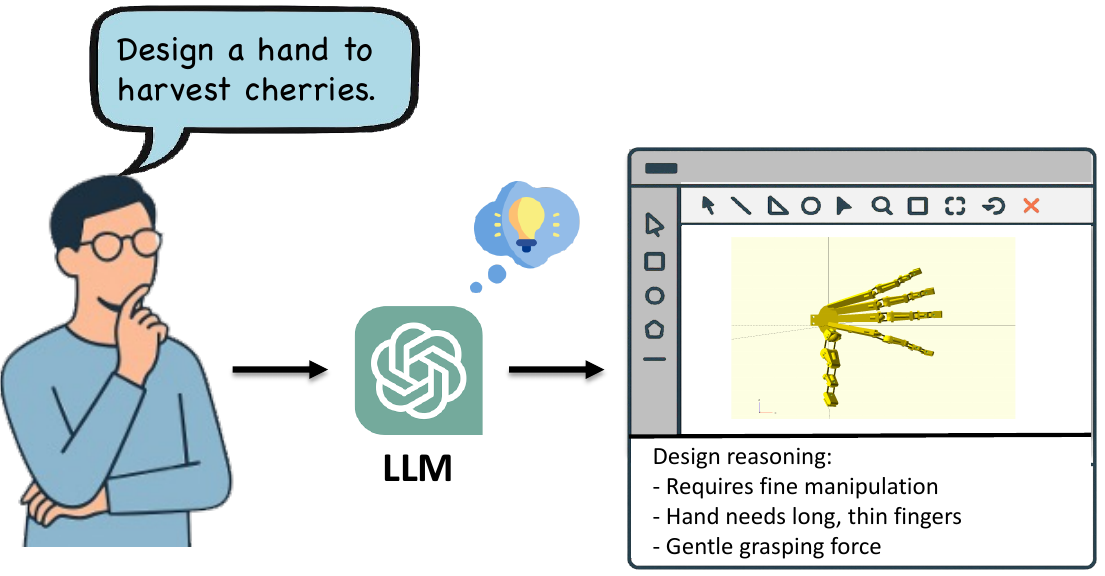}
   \vspace{-0pt}
      \caption{Lang2Morph transforms natural language task descriptions into robotic hand designs through LLM-guided semantic reasoning and structured generation. It analyzes task requirements such as user intent, grasp type, and object properties, and produces CAD-ready morphologies that align with functional goals and fabrication constraints.}
    \label{fig:teaser}
    \vspace{-12pt}
\end{figure}

In this paper, we propose \textbf{Lang2Morph}, a language-driven pipeline for robotic hand morphology generation. Our key insight is that LLMs are well suited to reasoning about task semantics and mapping them into symbolic and geometric design representations.  Lang2Morph combines this capability with structured validation and semantic feedback to generate task-conditioned morphologies directly from natural-language descriptions, as illustrated in Fig.~\ref{fig:teaser}, where a user-provided instruction is mapped into a design rationale and a CAD-ready morphology.

Lang2Morph builds upon the Open Parametric Hand (OPH) framework~\cite{gilday2025oph}, which defines a structured and fabricable design space for robotic hands. OPH supports single-piece 3D printing, allowing generated designs to be directly manufactured with minimal assembly. 
This enables our method to output physically realizable morphologies from natural-language instructions. Specifically, the pipeline comprises two major modules: (i) LLM-powered Morphology Design, which performs dual-level task analysis (semantic and structural), followed by geometry parameterization and constraint-aware validation to produce OPH-compatible parameters; and (ii) LLM-Guided Selection and Refinement, which ranks feasible rendered variants based on semantic alignment and size compatibility, while refinement is applied only when further correction is required. Together, these components form an end-to-end language-to-fabrication pipeline that generates task-adaptive robotic hand morphologies.

Our main contributions are as follows:
\begin{itemize}
    \item {We present Lang2Morph, an LLM-driven framework that generates fabricable, task-conditioned robotic hand morphologies directly from natural-language task descriptions within the OPH design space.}
    
    \item The pipeline consists of two key components: an LLM-powered morphology design module that progressively maps task semantics and structural requirements into detailed hand geometry, and an LLM-guided selection and refinement module that evaluates and improves generated candidates based on semantic alignment and size compatibility.
    
    \item We evaluate Lang2Morph through both morphology-level analysis and real-world cross-task experiments, demonstrating high design feasibility, morphological diversity, and distinct task-dependent physical capabilities in representative fabricated hands.
    
\end{itemize}

\section{RELATED WORK}

\noindent\textbf{Robot Hand Design.}
Robotic hand design has traditionally relied on expert engineering and task-specific heuristics. Representative platforms include underactuated and compliant hands such as Yale OpenHand~\cite{ma2013openhand}, iHY~\cite{odhner2013ihy}, Pisa/IIT SoftHand~\cite{catalano2014softhand}, RBO Hand 2~\cite{deimel2016rbohand2}, and DLR/HIT Hand II~\cite{liu2008dlrhit}. Adaptive and task-oriented robotic hands and grippers have also been developed through compliant structures, tunable geometries, and mechanically specialized designs~\cite{liu2023optimalgripper,chen2025clothhand,he2025robotic}. Recent open hardware systems, including BiDexHand~\cite{weng2025bidexhand} and ORCA~\cite{christoph2025orca}, further improve reproducibility and accessibility. Parametric frameworks such as the Open Parametric Hand (OPH)~\cite{gilday2025oph} expose structured, fabrication-aware design spaces for systematic morphology variation, but still largely rely on human designers to interpret task requirements and tune morphology parameters.

\noindent\textbf{Automated Robot Morphology Generation.}
Automated morphology generation has been studied through evolutionary search, grammar-based synthesis, and morphology--control co-optimization. Evolutionary approaches generate soft-bodied robots from voxel-based representations~\cite{cheney2013unshackling}, while RoboGrammar~\cite{robogrammar} uses graph grammars to construct robot topologies and Task2Morph~\cite{task2morph} jointly optimizes morphology and control in simulation. Related work has also jointly optimized robotic hand morphology and control for robust grasping across diverse objects~\cite{pan2021emergenthand}. 
Diversity-based topology optimization has further been explored to generate morphologically diverse soft robotic grippers with experimentally validated grasping capabilities \cite{pinskier2024diversity}. 
These approaches provide mechanisms for exploring robot design spaces, but typically rely on task-specific simulation objectives, optimization loops, or predefined search representations.

\noindent\textbf{LLM-Assisted Robot Design.}
Recent studies have begun to use large language models and vision-language models for robot morphology and embodiment design. Text2Robot~\cite{text2robot} generates manufacturable quadrupedal robots from textual specifications and combines morphology generation with control optimization. RoboMorph~\cite{qiu2024robomorph} integrates LLM reasoning with robot grammars and evolutionary feedback, while RoboMoRe~\cite{fang2025robomore} explores joint morphology and reward design. Foundation models have also been used for soft-robot evolution~\cite{chen2025naturalselection} and task-specific robotic tool generation~\cite{lin2025robotsmith,gao2025vlmgineer}, while Debate2Create~\cite{qiu2025debate2create} introduces multi-agent LLM debate for robot co-design. These works mainly address locomotion-oriented robots, soft or modular bodies, or external tools. In contrast, Lang2Morph focuses on task-conditioned robotic hand morphology generation directly from natural-language manipulation descriptions.

\begin{figure*}[t]
      \centering
      \includegraphics[width=0.8\textwidth]{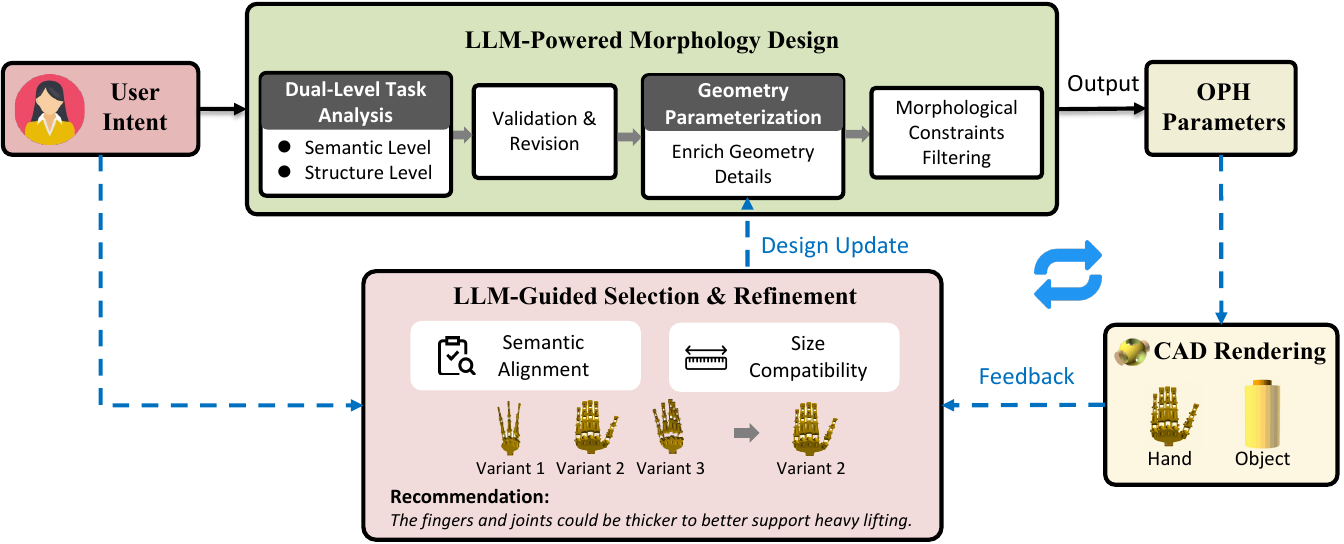}
       \caption{
Overview of the Lang2Morph pipeline for robotic hand morphology design.
Given a natural language user intent, an LLM extracts requirements and generates symbolic hand structures, which are validated, parameterized, and converted into OPH-compatible CAD designs.
Black arrows indicate the forward generation process; blue dashed arrows represent feedback-based refinement based on semantic alignment and size compatibility.
       }
      \label{fig:pipeline}
      \vspace{-0pt}
\end{figure*}

\section{METHODS}

\subsection{Overview}

We propose Lang2Morph, a language-driven framework for generating task-conditioned robotic hand morphologies from natural-language descriptions within the Open Parametric Hand (OPH) design space (Fig.~\ref{fig:pipeline}). The framework consists of two main stages. The first stage, LLM-powered morphology design, interprets the task at semantic and structural levels, generates detailed hand geometry, and applies deterministic validation to obtain feasible OPH-compatible designs. The second stage, LLM-guided selection and refinement, evaluates rendered candidates based on task alignment and size compatibility, selects the most suitable design, and provides refinement suggestions based on the evaluation feedback.

\subsection{OPH Design Space}
\label{sec:oph}
\begin{figure}[t]
   \centering
   \includegraphics[width=1\linewidth]{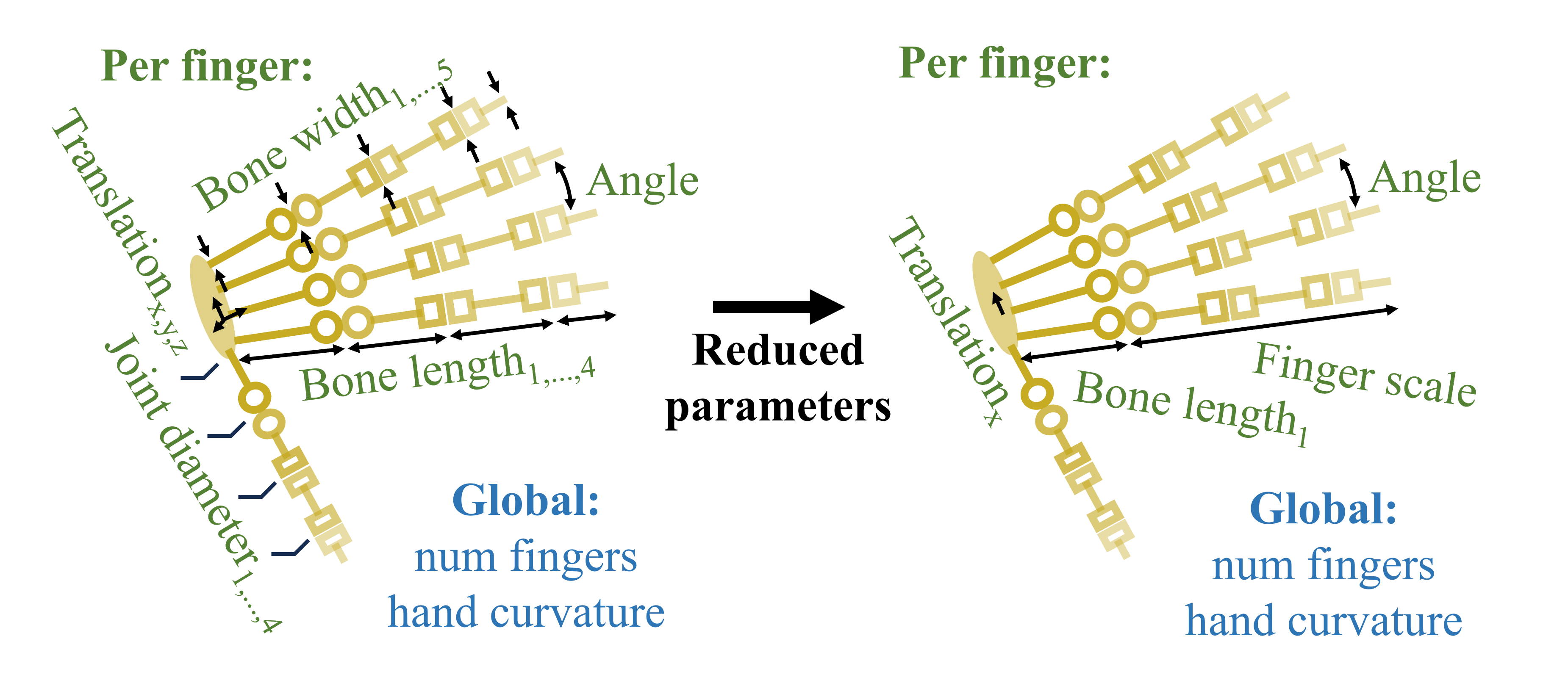}
   \vspace{-0pt}
\caption{Original OPH parameter space and reduced configuration used by Lang2Morph. 
}
    \label{fig:params}
    \vspace{-0pt}
\end{figure}

We adopt the Open Parametric Hand (OPH) framework as our base design space due to its compatibility with single-piece 3D printing and parametric control of morphology. 
The OPH defines up to 22 design parameters per finger plus number of fingers and a global hand curvature parameter. 
Specifically, for parameters with low impact on functional behaviors we use fixed or derived values, such as pulley and ligament thickness. For finger joints and bone lengths (except for the metacarpal), we reduce to one parameter (scale) per finger by fixing ratios according to the behaviourally optimized design in~\cite{carlet2025behaviour}. 
As illustrated in Fig.~\ref{fig:params}, we reduce the original OPH design space to a compact representation with task-dependent finger geometry and orientation parameters, together with two global parameters: finger count and hand curvature.
This compact representation preserves key functional differences while keeping the search space tractable.

\subsection{LLM-Powered Morphology Design}

The core of our pipeline is an LLM-powered design stage that maps natural-language task descriptions
into symbolic and geometric representations of robotic hands.

\subsubsection{Dual-Level Task Analysis}
The task description is analyzed at semantic and structural levels before geometry generation.

\noindent\textbf{Semantic Schema Generation.}
We begin by extracting task-implied functional requirements, as manipulation tasks differ in their force, precision, and contact demands.
For example, writing with a pen requires fine fingertip control, whereas lifting a heavy object emphasizes stable, load-bearing contact.
Accordingly, instead of directly generating geometry, we translate the task description into an explicit semantic schema that captures the required capabilities and target object context.

We use a large language model (LLM) to map the natural-language intent $T$ to a structured semantic schema:
\begin{equation}
\mathcal{F} = f_{\mathrm{LLM}}(T),
\end{equation}
where $\mathcal{F}$ encodes task-relevant information for downstream structure and parameter generation. The schema includes:

\begin{itemize}
    \item \textit{Task understanding:} identify the primary goal, the object being manipulated, and the required levels of force and precision.
    \item \textit{Object property extraction:} estimate physical properties such as size, material, fragility, and surface friction.
    \item \textit{Grasp type classification:} categorize the task into a representative grasp type.
\end{itemize}

Among the extracted attributes, grasp type plays a central role in guiding design decisions. To provide an interpretable intermediate representation and improve consistency across task descriptions, we adopt a simplified taxonomy that maps each task to one of three grasp types: \emph{force-based}, \emph{fine manipulation}, or \emph{tool-based}. This abstraction is inspired by established human and robotic grasp taxonomies~\cite{cutkosky1989grasp, feix2016grasp, kivell2015evidence} and captures dominant functional requirements in a compact form.
In Lang2Morph, grasp type is used as an intermediate semantic signal that conditions downstream structure generation and parameterization, improving consistency across diverse natural-language instructions.

\noindent\textbf{Structure Grammar Generation.}
To represent the hand’s structure, we adopt a graph grammar formalism inspired by RoboGrammar~\cite{robogrammar}. The grammar is defined as:
\begin{equation}
\mathcal{G}_{\mathrm{hand}} = (N, T, A, R, S),
\end{equation}
where $N$ and $T$ are nonterminal and terminal symbols, $A$ is a set of symbolic attributes, $R$ is the set of production rules, and $S$ is the start symbol. Components include P (palm), F (finger), J (joint), L (link), T (tendon), M (mount), and C (connector).
Grammar generation is guided by the semantic tags produced in the previous step. 
Specifically,
the LLM is prompted to output a JSON-style grammar that includes:

\begin{itemize}
    \item Component definitions: symbolic identifiers for palm, finger, joint, etc.
    \item Structure rules: hierarchical expansions using bidirectional and sequential connections.
    \item Connection rules: how fingers attach to mounts and the palm.
    \item Spatial layout and actuation hints: layout priors such as finger spacing and tendon routing preferences.
\end{itemize}

An example rule set for a symmetric hand includes:
\begin{center}
\ttfamily
S $\rightarrow$ P $\leftrightarrow$ F1 $\leftrightarrow$ F2 $\leftrightarrow$ F3 \\
F1 $\rightarrow$ J1 $\leftrightarrow$ L1 $\leftrightarrow$ J2 $\leftrightarrow$ L2 $\leftrightarrow$ J3 \\
F2 $\rightarrow$ J4 $\leftrightarrow$ L3 $\leftrightarrow$ J5 $\leftrightarrow$ L4 $\leftrightarrow$ J6 \\
F3 $\rightarrow$ J7 $\leftrightarrow$ L5 $\leftrightarrow$ J8 $\leftrightarrow$ L6 $\leftrightarrow$ J9
\end{center}

These grammar rules define the high-level topology of the robotic hand, providing an intermediate structural representation that organizes component connectivity and finger configurations before geometric parameterization. This representation also enables structural consistency checks before generating OPH-compatible parameters.

\subsubsection{Automatic Validation and Revision.}
Before geometry generation, the generated structural grammar is passed through a hybrid validation stage. 
First, programmatic checks verify basic structural completeness and consistency, including the presence of valid structure and connection rules, finger symbols/components, and consistent connectivity. 
An LLM-based structural validator then provides complementary diagnostic feedback on structural plausibility and potential design issues.
The deterministic validation result serves as the primary gate for downstream generation, while the LLM assessment is used to provide revision guidance. 
If the grammar does not satisfy the structural requirements, the validation feedback is incorporated into a constrained revision of the grammar, followed by revalidation before geometry parameterization. 
Candidates that remain invalid are excluded from downstream generation.

\subsubsection{Geometry Parameterization.}
Conditioned on the validated structural grammar $S^*$ and semantic representation $\mathcal{F}$, the LLM generates a structured geometric parameter set $\theta$:
\begin{equation}
\theta = f_{\mathrm{LLM}}(S^*, \mathcal{F}).
\end{equation}

The OPH-derived representation includes global variables for finger number and hand curvature, together with per-finger parameters such as finger length, joint and width dimensions, and orientation. Lower-level fabrication parameters and finger root positions are fixed or deterministically derived from the CAD template. The generated parameters are mapped to OpenSCAD and subsequently checked by the morphological constraints described below.
Grasp-category priors and variant-specific cues further guide geometry generation toward task-relevant designs, e.g., slender fingers for fine manipulation, more robust structures for force-oriented tasks, and suitable opposition/orientation for tool use.

\subsubsection{Morphological Constraint Filtering.}
\label{subsubsec:morphological_constraint_filter}
Before converting the generated geometry representation into a fabrication-ready CAD model, we apply deterministic morphological screening at both the structured-geometry and CAD-parameter levels. The geometry-level filter checks whether the generated representation is complete and internally consistent, including required finger-level parameters, segment-length ratios, and global hand properties. After conversion to the OPH-compatible CAD representation, additional constraints are imposed on joint and link dimensions, segment slenderness, total finger length, finger orientation, and the approximate fabrication footprint.
For example, joint diameters and link widths are restricted to predefined geometric ranges, the total finger length is limited to $200\,\mathrm{mm}$, and finger orientations are bounded to avoid implausible initial configurations. Minor numerical violations in selected geometric parameters may be softly adjusted before final validation, whereas candidates with unresolved structural or geometric violations are excluded from downstream rendering and evaluation. These constraints provide deterministic geometric and fabrication-plausibility screening rather than a physical simulation or guarantee of mechanical stability.

\subsubsection{Diversity Strategy.}
To encourage that variants generated for the same task are structurally distinct, we introduce a diversity strategy during grammar generation. 
Instead of relying purely on stochastic sampling, we prompt the LLM with variant-specific design cues that bias the generation toward different structural layouts (e.g., symmetric vs.\ asymmetric, compact vs.\ extended). 
For example, a prompt may include: 
\emph{``Use a manufacturing-friendly design optimized for 3D printing constraints, favoring modular and symmetric structures.''} 
Such cues encourage the LLM to produce distinct morphologies while preserving the same functional requirements. 
This mechanism underpins the multi-variant design process, providing a richer candidate pool for downstream ranking and feedback.

\begin{figure}[t]
\centering
\includegraphics[width=\linewidth]{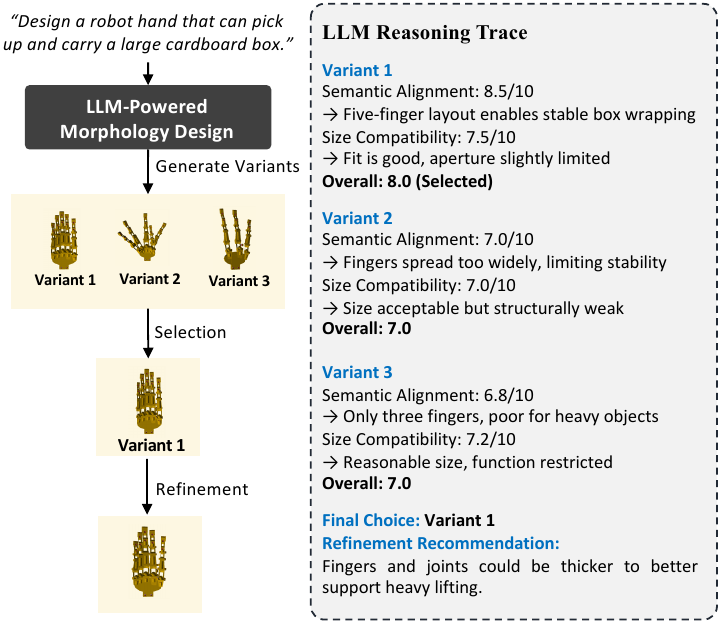}
\caption{%
{Illustration of LLM-guided selection and refinement. Feasible hand variants are evaluated for semantic alignment and size compatibility, ranked, and the top candidate is selected. The right panel shows a simplified evaluation trace and refinement feedback.}
}
\label{fig:rank}
\vspace{-0pt}
\end{figure}

\subsection{LLM-Guided Selection and Refinement}

After generating feasible hand variants using the LLM-based design pipeline, we introduce a feedback module to evaluate and rank the candidates, as illustrated in Fig.~\ref{fig:rank}. The top-ranked design is selected as the final candidate and can be further refined based on the evaluation feedback when necessary.

\noindent\textbf{Ranking and Selection.}
The selection module evaluates each feasible design according to two complementary criteria: semantic alignment with the intended task and size compatibility with the target object.

\begin{itemize}
    \item \textit{Semantic Alignment}: An LLM assesses whether each design variant is consistent with the original task description. The evaluation prompt includes the user intent, a summary of the design rationale, geometric parameters, and descriptions of the rendered hand generated from multi-view images by a multimodal model. The LLM assigns a task-alignment score from 0 to 10 together with a short justification.

    \item \textit{Size Compatibility}: The same multi-view renderings are used to describe the relative scale and spatial relationship between the generated hand and the target object. These descriptions, together with the task context and geometric parameters, are evaluated by the LLM to produce a size-compatibility score from 0 to 10.
\end{itemize}

The two scores are averaged to obtain an overall evaluation score. Feasible candidates are ranked accordingly, and the highest-scoring design is selected for downstream use.

\noindent\textbf{Refinement.}
Based on the evaluation results, the LLM determines whether the selected design requires further refinement. If improvements are suggested, the feedback is translated into parameter-level adjustments and incorporated into a revised generation prompt. Otherwise, the selected design is retained without further modification.

\section{EXPERIMENTS}
\label{sec:experiments}

\subsection{Experimental Setup}

\noindent\textbf{Task Set and Implementation.}
We evaluate Lang2Morph on 30 natural-language task descriptions spanning three grasp categories: force-based, fine manipulation, and tool-based. For each task, the pipeline generates three design variants, resulting in 90 requested designs. We use GPT-5.4-mini as the LLM backbone for both Lang2Morph and the Direct LLM baseline. Generated designs are represented in structured JSON and converted to OPH-compatible OpenSCAD models for downstream evaluation and fabrication.

\noindent\textbf{Baselines.}
We compare Lang2Morph with three baselines. \textit{Random }samples OPH parameters from broad engineering ranges without task information. \textit{Template heuristic} receives expert-labeled task attributes and applies hand-written template rules with bounded parameter sweeps; it serves as a hand-engineered feasibility reference rather than an automated language-to-design method. \textit{Direct LLM} uses the same LLM model to directly predict OPH parameters from the task description without Lang2Morph's structured reasoning process. All methods share the same OPH parameterization and feasibility checks; only the template heuristic uses manually specified task-to-template mappings.

\begin{table}[t]
\centering
\caption{Design feasibility comparison across generation methods.}
\label{tab:method_baseline_comparison}
\resizebox{\columnwidth}{!}{
\begin{tabular}{lccc}
\toprule
\textbf{Method} & \textbf{Input} & \textbf{Manual Mapping} & \textbf{DFR (\%)} 
\\
\midrule
Random & None & None & 28.9 \\
Template heuristic & Expert attrs. & Templates & 100.0 \\
Direct LLM & Language & None & 30.0 \\
Lang2Morph & Language & None & 97.8 \\
\bottomrule
\end{tabular}
}
\vspace{-0pt}
\end{table}

\subsection{Design Feasibility Evaluation}

\noindent\textbf{Evaluation Metric.}
To evaluate whether generated designs satisfy structural and geometric constraints and are suitable for downstream fabrication, we compute the Design Feasibility Rate (DFR):
\begin{equation}
\text{DFR} = \frac{N_{\text{feasible}}}{N_{\text{requested}}},
\end{equation}
where $N_{\text{feasible}}$ denotes the number of generated variants that satisfy the morphological constraints and can be instantiated as valid OPH-compatible designs, and $N_{\text{requested}}$ is the total number of requested variants. The feasibility checks cover finger count, joint and link dimensions, slenderness ratios, finger length, initial orientations, and fabrication footprint.

\noindent\textbf{Results.}
As shown in Table~\ref{tab:method_baseline_comparison}, Lang2Morph achieves a DFR of 97.8\%, approaching the 100.0\% feasibility of the expert-defined template baseline while operating directly from natural-language task descriptions. In contrast, Random and Direct LLM achieve only 28.9\% and 30.0\% DFR, respectively, indicating that unconstrained sampling or direct language-to-geometry prediction frequently produces designs that violate OPH feasibility constraints. The large gap between Direct LLM and Lang2Morph suggests that the structured generation and validation mechanisms substantially improve design feasibility over direct parameter prediction.

\subsection{Morphology Design Space: Coverage and Separation}

\noindent\textbf{Evaluation Metrics.}
To analyze how different design methods organize the morphology space, we represent each generated hand using a set of continuous morphology descriptors capturing global hand shape, finger dimensions, relative joint size, and finger orientations and spatial arrangement. We then apply Principal Component Analysis (PCA)~\cite{jolliffe2016pca} to visualize the resulting morphology distributions in a two-dimensional space. The descriptors are jointly standardized across all methods, and a single PCA model is fitted to the pooled feature vectors.

\begin{figure}[t]
\centering
\includegraphics[width=\linewidth]{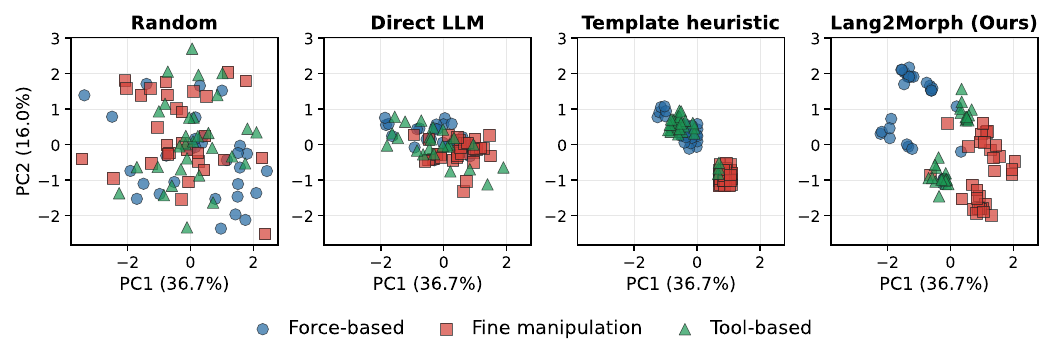}
\caption{PCA visualization of morphology distributions generated by different methods across the three grasp categories.}
\label{fig:design-space-pca}
\vspace{-0pt}
\end{figure}

\noindent\textbf{Results.}
Fig.~\ref{fig:design-space-pca} compares the morphology distributions produced by \emph{Random}, \emph{Direct LLM}, \emph{Template heuristic}, and \emph{Lang2Morph}. \emph{Random} produces a broadly dispersed but strongly intermixed distribution, showing little correspondence between morphology variation and task category. \emph{Direct LLM} exhibits partial task-dependent organization, but substantial overlap remains among the three grasp categories.
The \emph{Template heuristic} produces compact, category-biased regions determined by predefined task-to-template mappings. Fine-manipulation designs form a relatively distinct group, whereas force-based and tool-based designs occupy nearby and partially overlapping regions. The limited spatial coverage and low within-category dispersion are consistent with the restricted variation of the predefined templates.
In contrast, \emph{Lang2Morph} produces clearly differentiated task-conditioned regions while retaining substantial within-category variation. Fine-manipulation, force-based, and tool-based designs occupy different portions of the projected morphology space, indicating systematic morphology differences across task categories. The visible dispersion within each category further suggests that Lang2Morph generates diverse morphology candidates rather than repeatedly reproducing a single task-specific template.
The quantitative analysis is consistent with the PCA visualization. \emph{Random} and \emph{Direct LLM} show little category separation, with silhouette scores of -0.020 and 0.002, respectively. The \emph{Template heuristic} achieves the strongest category separation (0.382), but also exhibits the lowest within-category spread (0.543). In comparison, \emph{Lang2Morph} achieves a positive silhouette score of 0.221 while maintaining substantially greater within-category variation (1.475), indicating a balance between task-dependent morphological organization and design diversity.

\begin{figure}[t]
  \centering
  \includegraphics[width=0.89\linewidth]{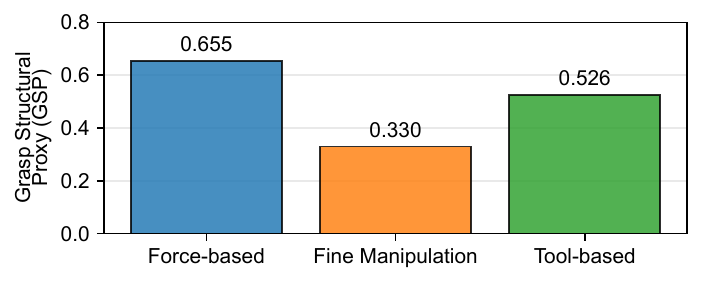}
  \vspace{-0pt}
\caption{
    Average Grasp Structural Proxy (GSP) across different task categories. GSP is a morphology-based descriptor computed from five normalized structural factors.
}
  \label{fig:grasp-capability}
\vspace{-0pt}
\end{figure}

\begin{figure*}[t]
      \centering
      \includegraphics[width=\textwidth]{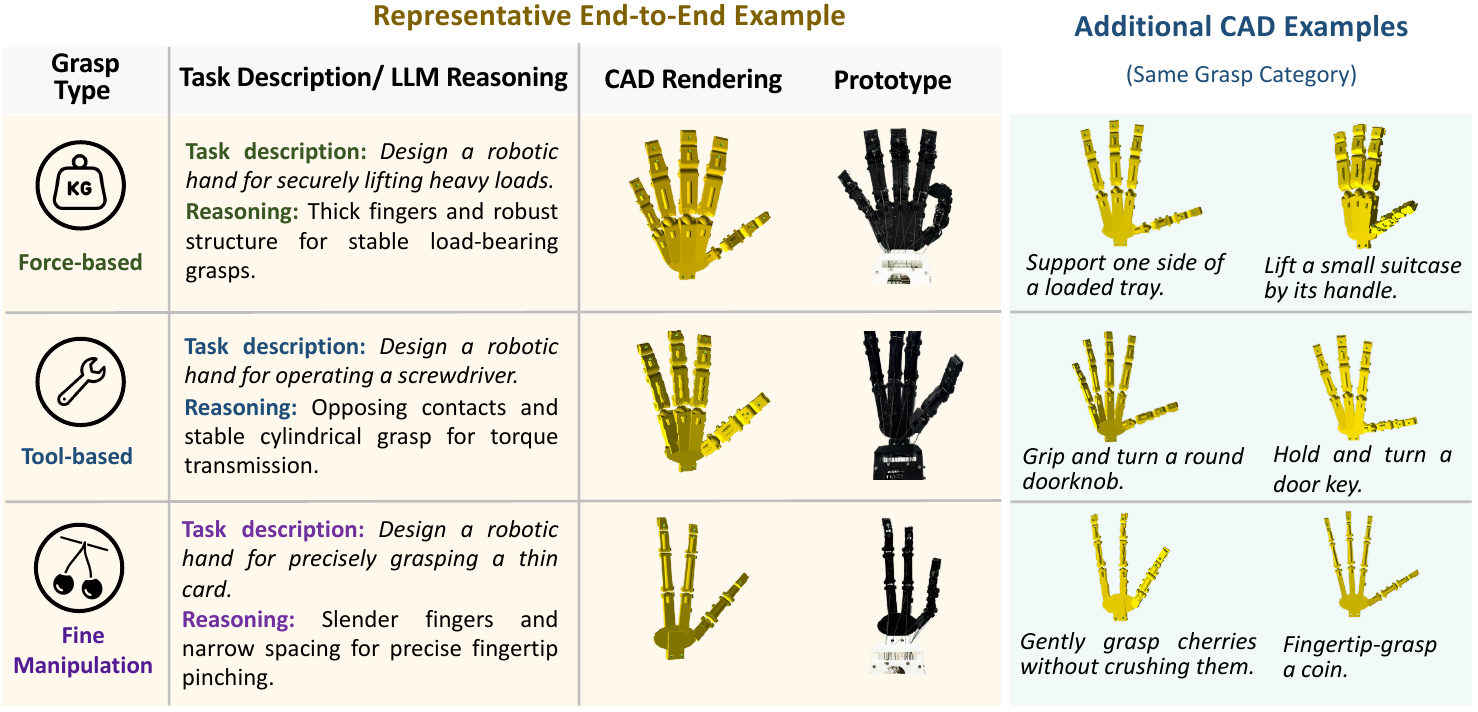}
\caption{Representative and additional hand designs generated by Lang2Morph. For each grasp category, the left panel presents an end-to-end example linking the task description and LLM design rationale to the CAD morphology and fabricated prototype. The right panel shows additional CAD examples from the same grasp category, illustrating intra-category variation rather than collapse to a single template.}
      \label{fig:qualitative}
      \vspace{-0pt}
\end{figure*}

\begin{table}[t]
\centering
\caption{Run-to-run Design Feasibility Rate (DFR) on a stratified 10-task subset.}
\label{tab:run_stability}
\begin{tabular}{lcc}
\toprule
\textbf{Run} & \textbf{Feasible Designs} & \textbf{DFR (\%)} \\
\midrule
Run 1 & 29/30 & 96.7 \\
Run 2 & 29/30 & 96.7 \\
Run 3 & 27/30 & 90.0 \\
\midrule
Mean $\pm$ std. & -- & $94.4 \pm 3.8$ \\
\bottomrule
\end{tabular}
\vspace{-0pt}
\end{table}

\subsection{Morphology-Based Structural Characteristics}

\noindent\textbf{Evaluation Metric.}
To examine whether the generated morphologies exhibit systematic category-dependent structural differences, we conduct a morphology-level analysis over the full set of generated designs. Specifically, we compute a morphology-based Grasp Structural Proxy (GSP) from normalized morphology parameters to summarize grasp-related structural characteristics. GSP is intended as a comparative structural descriptor rather than a direct measurement of physical grasping force. Specifically, GSP integrates five normalized factors: average joint diameter ($f_{\mathrm{joint}}$), finger width ($f_{\mathrm{width}}$), finger length ($f_{\mathrm{length}}$), number of enabled fingers ($f_{\mathrm{count}}$), and intra-hand morphological consistency ($f_{\mathrm{harmony}}$). Each component is scaled to $[0,1]$ and combined using fixed heuristic weights, with larger weights assigned to joint diameter and finger width. The resulting score ranges from 0 to 1 and is used for relative comparison across generated morphologies.

\noindent\textbf{Results.}
As shown in Fig.~\ref{fig:grasp-capability}, the average GSP differs across task categories. Force-based tasks obtain the highest score, reflecting morphologies with more robust grasp-related structural characteristics. Tool-based tasks also obtain a relatively high score, consistent with the need for stable and supportive contact during tool manipulation. Fine-manipulation tasks produce the lowest score, reflecting their greater emphasis on slender structures and precise contact.
Overall, these results show systematic category-dependent differences in grasp-related structural characteristics across the generated morphologies. Lang2Morph adapts properties such as finger thickness, length, and finger count according to the functional requirements inferred from task descriptions.

\subsection{Run-to-Run Stability}

To examine the run-to-run stability of the LLM-based generation process, we select a stratified subset of 10 tasks from the full 30-task evaluation set, covering all three grasp categories. The subset is chosen to provide representative coverage of force-based, fine-manipulation, and tool-based tasks while keeping repeated generation computationally manageable. For each task, the same three variant conditions are generated in three independent runs under identical model, prompt, validation, and CAD settings.
As shown in Table~\ref{tab:run_stability}, the three runs achieve DFRs of 96.7\%, 96.7\%, and 90.0\%, with a mean of 94.4\% $\pm$ 3.8\%. Grasp-category predictions remain fully consistent across repetitions after label normalization. For matched task--variant outputs, we compute an RMS distance over seven normalized morphology descriptors capturing hand curvature, relative finger dimensions, finger orientations, and spatial arrangement. The mean RMS distance is 0.0517, indicating relatively small geometric variation across repeated generations.

\begin{figure*}[t]
    \centering
    \includegraphics[width=0.82\textwidth]{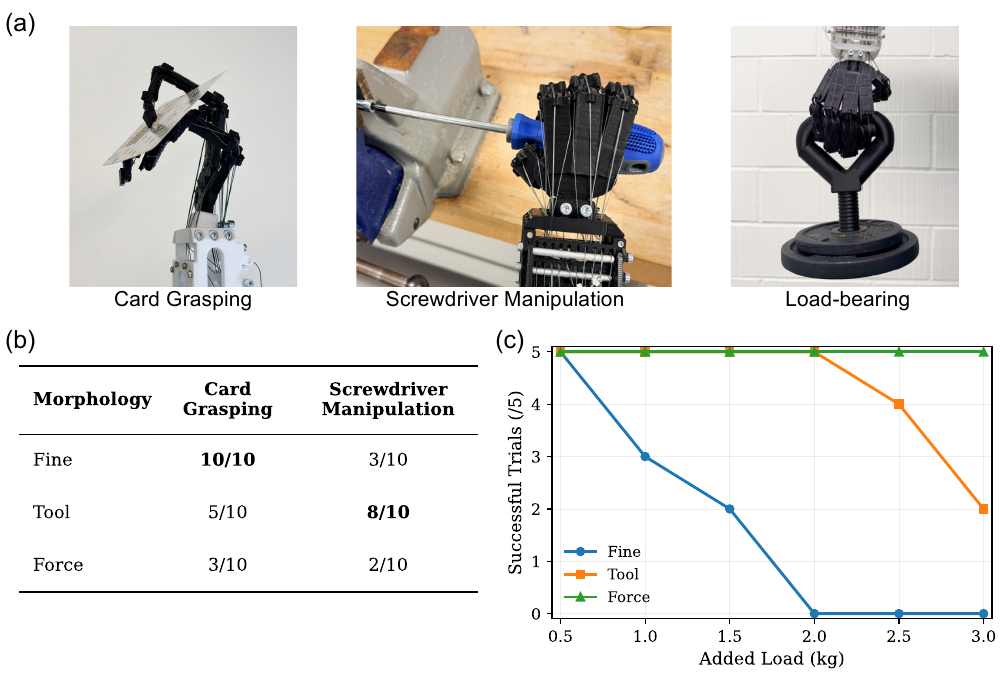}
\caption{
Real-world cross-task evaluation of three fabricated robotic hand morphologies. (a) Representative demonstrations of card grasping, screwdriver manipulation, and load-bearing tasks. All three hands were evaluated on the same three tasks. (b) Success rates for card grasping and screwdriver manipulation, each evaluated over 10 trials per hand. (c) Load-bearing performance under progressively increasing added loads from 0.5 to 3.0 kg, with five trials per load condition.
}
    \label{fig:cross-task-real}
\end{figure*}

\subsection{{Real-World Functional Validation}}

We fabricate one representative feasible design from each grasp category, referred to as the \emph{Fine-manipulation}, \emph{Tool-based}, and \emph{Force-based} hands. The designs are primarily selected based on the LLM-guided ranking, with practical considerations such as printability and mounting compatibility used to resolve selections among similarly scored candidates. All three hands are evaluated across the same physical tasks to examine whether their task-conditioned morphological differences lead to distinct functional characteristics. Fig.~\ref{fig:qualitative} summarizes representative end-to-end examples and additional generated morphologies within each category.

\noindent\textbf{Experimental Protocol.}
The evaluation considers three complementary aspects of hand functionality. Card grasping emphasizes precise contact and stable manipulation of a thin object, screwdriver manipulation requires stable tool enclosure and torque transmission, and load-bearing evaluates the ability to maintain a grasp under increasing external load. Card grasping and screwdriver manipulation are each repeated for 10 trials per hand. Card trials require stable grasping and transport without dropping, while screwdriver trials require three effective rotations with observable screw advancement.
For the load-bearing evaluation, each hand grasps the same 158-g printed handle while the added load is progressively increased from 0.5 to 3.0~kg in 0.5-kg increments, with five trials per load condition. The progressive loading protocol characterizes the load-bearing profile of each morphology across increasing load demands. A trial succeeds if the hand lifts and maintains the suspended load for 5~s. 
Representative experimental setups for the three tasks are shown in Fig.~\ref{fig:cross-task-real}(a).

\noindent\textbf{Cross-task Results.}
As shown in Fig.~\ref{fig:cross-task-real}(b), the three fabricated morphologies exhibit distinct but overlapping functional characteristics. The Fine-manipulation hand achieves the highest success in card grasping, while the Tool-based hand performs best in screwdriver manipulation. The Force-based hand performs less effectively on these two manipulation tasks. The non-specialized hands nevertheless retain partial cross-task capability, indicating functional preferences rather than rigid task-exclusive behaviors.
Fig.~\ref{fig:cross-task-real}(c) further reveals different load-bearing profiles under increasing load demands. The Force-based hand succeeds in all trials throughout the tested range up to 3.0~kg, with its capacity limit not reached within this range. The Tool-based hand remains fully successful through 2.0~kg before performance decreases at higher loads. In contrast, the Fine-manipulation hand degrades progressively as the load increases and fails all trials from 2.0~kg onward.
This trend is qualitatively consistent with the category-level structural differences indicated by GSP in Fig.~\ref{fig:grasp-capability}.
Overall, the three hands exhibit task-dependent performance differences: the Fine-manipulation hand performs best in card grasping, the Tool-based hand in screwdriver manipulation, and the Force-based hand under higher loads. These results demonstrate distinct functional preferences among the three fabricated task-conditioned morphologies.

\noindent\textbf{Failure Modes.}
In screwdriver manipulation, the Fine-manipulation hand primarily loses the grasp as the screwdriver slips downward, whereas the screwdriver can rotate within the larger central opening of the Force-based hand without producing effective screw rotation. In card grasping, failures mainly arise from unsuccessful initial contact or dropping during transport. In the load-bearing tests, observed failures primarily involve slippage or inability to sustain the applied load.

\section{CONCLUSION}

We presented Lang2Morph, a language-driven framework for generating fabricable, task-conditioned robotic hand morphologies from natural-language descriptions. By combining LLM-based task reasoning, structured morphology generation, geometric constraints, and feedback-guided selection, 
Lang2Morph achieves a design feasibility rate comparable to expert-defined template methods while producing diverse task-conditioned morphologies. 
Real-world cross-task experiments further show that the generated hands exhibit distinct physical capabilities across force-based, tool-based, and fine-manipulation tasks. These results suggest that language models can serve as practical interfaces for robotic morphology design, with the potential to scale across diverse task requirements without requiring a manually specified task-to-template mapping or a task-specific optimization objective for each input task. Future work will consider broader manipulation settings and more systematic mechanical evaluation.

\bibliography{IEEEexample}

\begin{thebibliography}{32}
\providecommand{\natexlab}[1]{#1}
\providecommand{\url}[1]{\texttt{#1}}
\expandafter\ifx\csname urlstyle\endcsname\relax
  \providecommand{\doi}[1]{doi: #1}\else
  \providecommand{\doi}{doi: \begingroup \urlstyle{rm}\Url}\fi

\bibitem[Odhner et~al.(2013)Odhner, Jentoft, Claffee, Corson, Tenzer, Ma,
  Buehler, Kohout, Howe, and Dollar]{odhner2013ihy}
Lael Odhner, Leif~P. Jentoft, Mark~R. Claffee, Nick Corson, Yaroslav Tenzer,
  Raymond~R. Ma, Martin Buehler, Robert~C. Kohout, Robert~D. Howe, and Aaron~M.
  Dollar.
\newblock A compliant, underactuated hand for robust manipulation.
\newblock \emph{The International Journal of Robotics Research}, 33:\penalty0
  736 -- 752, 2013.

\bibitem[Liu et~al.(2008)Liu, Wu, Meusel, Seitz, Hirzinger, Jin, Liu, Fan, Lan,
  and Chen]{liu2008dlrhit}
Hong Liu, Ke~Wu, Peter Meusel, Nikolaus Seitz, Gerd Hirzinger, MH~Jin, YW~Liu,
  SW~Fan, T~Lan, and ZP~Chen.
\newblock Multisensory five-finger dexterous hand: The dlr/hit hand ii.
\newblock In \emph{Proc. IEEE/RSJ Int. Conf. on Intelligent Robots and
  Systems}, 2008.

\bibitem[Ma et~al.(2013)Ma, Odhner, and Dollar]{ma2013openhand}
Raymond~R. Ma, Lael~U. Odhner, and Aaron~M. Dollar.
\newblock A modular, open-source 3d printed underactuated hand.
\newblock In \emph{Proc. IEEE Int. Conf. on Robotics and Automation}, pages
  2737--2743, 2013.

\bibitem[Ma and Dollar(2017)]{ma2017openhand2}
Raymond~R. Ma and Aaron~M. Dollar.
\newblock Yale openhand project: Optimizing open-source hand designs for ease
  of fabrication and adoption.
\newblock \emph{IEEE Robotics \& Automation Magazine}, 24\penalty0
  (1):\penalty0 32--40, 2017.

\bibitem[Gilday et~al.(2025)Gilday, Sirithunge, Iida, and
  Hughes]{gilday2025oph}
Kieran Gilday, Chapa Sirithunge, Fumiya Iida, and Josie Hughes.
\newblock Embodied manipulation with past and future morphologies through an
  open parametric hand design.
\newblock \emph{Science Robotics}, 2025.
\newblock \doi{10.1126/scirobotics.ads6437}.

\bibitem[Cheney et~al.(2013)Cheney, MacCurdy, Clune, and
  Lipson]{cheney2013unshackling}
Nick Cheney, Robert MacCurdy, Jeff Clune, and Hod Lipson.
\newblock Unshackling evolution: evolving soft robots with multiple materials
  and a powerful generative encoding.
\newblock In \emph{Proc. Genetic and Evolutionary Computation Conference},
  pages 167--174. ACM, 2013.
\newblock \doi{10.1145/2463372.2463404}.

\bibitem[Zhao et~al.(2020)Zhao, Xu, Konakovic-Lukovic, Hughes, Spielberg, Rus,
  and Matusik]{robogrammar}
Allan Zhao, Jie Xu, Mina Konakovic-Lukovic, Josephine Hughes, Andrew Spielberg,
  Daniela Rus, and Wojciech Matusik.
\newblock Robogrammar: Graph grammar for terrain-optimized robot design.
\newblock \emph{ACM Trans. Graph. (SIGGRAPH)}, 39\penalty0 (6):\penalty0
  188:1--188:16, 2020.
\newblock \doi{10.1145/3414685.3417831}.

\bibitem[Cai et~al.(2023)Cai, Yang, Li, Chen, Mao, Yi, and Yang]{task2morph}
Yishuai Cai, Shaowu Yang, Minglong Li, Xinglin Chen, Yunxin Mao, Xiaodong Yi,
  and Wenjing Yang.
\newblock Task2morph: Differentiable task-inspired framework for contact-aware
  robot design.
\newblock In \emph{Proc. IEEE/RSJ International Conference on Intelligent
  Robots and Systems (IROS)}, pages 452--459, 2023.
\newblock \doi{10.1109/IROS55552.2023.10342117}.

\bibitem[Ahn et~al.(2022)Ahn, Brohan, Brown, Chebotar, Cortes, David, Finn,
  Gopalakrishnan, Hausman, Herzog, Ho, Hsu, Ibarz, Ichter, Irpan, Jang, Ruano,
  Jeffrey, Jesmonth, Joshi, Julian, Kalashnikov, Kuang, Lee, Levine, Lu, Luu,
  Parada, Pastor, Quiambao, Rao, Rettinghouse, Reyes, Sermanet, Sievers, Tan,
  Toshev, Vanhoucke, Xia, Xiao, Xu, Xu, and Yan]{ahn2022saycan}
Michael Ahn, Anthony Brohan, Noah Brown, Yevgen Chebotar, Omar Cortes, Byron
  David, Chelsea Finn, Keerthana Gopalakrishnan, Karol Hausman, Alexander
  Herzog, Daniel Ho, Jasmine Hsu, Julian Ibarz, Brian Ichter, Alex Irpan, Eric
  Jang, Rosario M~Jauregui Ruano, Kyle Jeffrey, Sally Jesmonth, Nikhil~Jayant
  Joshi, Ryan~C. Julian, Dmitry Kalashnikov, Yuheng Kuang, Kuang-Huei Lee,
  Sergey Levine, Yao Lu, Linda Luu, Carolina Parada, Peter Pastor, Jornell
  Quiambao, Kanishka Rao, Jarek Rettinghouse, Diego~M Reyes, Pierre Sermanet,
  Nicolas Sievers, Clayton Tan, Alexander Toshev, Vincent Vanhoucke, F.~Xia,
  Ted Xiao, Peng Xu, Sichun Xu, and Mengyuan Yan.
\newblock Do as i can, not as i say: Grounding language in robotic affordances.
\newblock In \emph{Conference on Robot Learning}, 2022.

\bibitem[Zhou et~al.(2023)Zhou, Hong, and Wu]{Zhou2023NavGPTER}
Gengze Zhou, Yicong Hong, and Qi~Wu.
\newblock Navgpt: Explicit reasoning in vision-and-language navigation with
  large language models.
\newblock In \emph{AAAI Conference on Artificial Intelligence}, 2023.

\bibitem[Driess et~al.(2023)Driess, Xia, Sajjadi, Lynch, Chowdhery, Ichter,
  Wahid, Tompson, Vuong, Yu, Huang, Chebotar, Sermanet, Duckworth, Levine,
  Vanhoucke, Hausman, Toussaint, Greff, Zeng, Mordatch, and
  Florence]{driess2023palme}
Danny Driess, F.~Xia, Mehdi S.~M. Sajjadi, Corey Lynch, Aakanksha Chowdhery,
  Brian Ichter, Ayzaan Wahid, Jonathan Tompson, Quan~Ho Vuong, Tianhe Yu,
  Wenlong Huang, Yevgen Chebotar, Pierre Sermanet, Daniel Duckworth, Sergey
  Levine, Vincent Vanhoucke, Karol Hausman, Marc Toussaint, Klaus Greff, Andy
  Zeng, Igor Mordatch, and Peter~R. Florence.
\newblock Palm-e: An embodied multimodal language model.
\newblock In \emph{International Conference on Machine Learning}, 2023.

\bibitem[Ringel et~al.(2025)Ringel, Charlick, Liu, Xia, and Chen]{text2robot}
Ryan~P Ringel, Zachary~S Charlick, Jiaxun Liu, Boxi Xia, and Boyuan Chen.
\newblock Text2robot: Evolutionary robot design from text descriptions.
\newblock In \emph{IEEE International Conference on Robotics and Automation
  (ICRA)}, pages 5789--5797. IEEE, 2025.

\bibitem[Fang et~al.(2025)Fang, Sun, Ma, Lu, and Yao]{fang2025robomore}
Jiawei Fang, Yuxuan Sun, Chengtian Ma, Qiuyu Lu, and Lining Yao.
\newblock Robomore: Llm-based robot co-design via joint optimization of
  morphology and reward.
\newblock \emph{arXiv preprint arXiv:2506.00276}, 2025.

\bibitem[Lin et~al.(2025)Lin, Yuan, Wang, Qiu, Wang, Guo, Wang, Narang, Fox,
  and Gan]{lin2025robotsmith}
Chunru Lin, Haotian Yuan, Yian Wang, Xiaowen Qiu, Tsun-Hsuan~Johnson Wang,
  Minghao Guo, Bohan Wang, Yashraj Narang, Dieter Fox, and Chuang Gan.
\newblock Robotsmith: Generative robotic tool design for acquisition of complex
  manipulation skills.
\newblock In \emph{Advances in Neural Information Processing Systems},
  volume~38, 2025.

\bibitem[Catalano et~al.(2014)Catalano, Grioli, Farnioli, Serio, Piazza, and
  Bicchi]{catalano2014softhand}
Manuel~G. Catalano, Giorgio Grioli, Emanuele Farnioli, Andrea Serio, Carlo
  Piazza, and Antonio Bicchi.
\newblock Adaptive synergies for the design and control of the pisa/iit
  softhand.
\newblock \emph{The International Journal of Robotics Research}, 2014.
\newblock \doi{10.1177/0278364913518998}.

\bibitem[Deimel and Brock(2016)]{deimel2016rbohand2}
Raphael Deimel and Oliver Brock.
\newblock A novel type of compliant, underactuated robotic hand for dexterous
  grasping.
\newblock \emph{The International Journal of Robotics Research}, 2016.
\newblock \doi{10.1177/0278364915592961}.

\bibitem[Liu et~al.(2023)Liu, Yang, and Shih]{liu2023optimalgripper}
Chih-Hsing Liu, Sy-Yeu Yang, and Yi-Chieh Shih.
\newblock Optimal design of a highly self-adaptive gripper with multi-phalange
  compliant fingers for grasping irregularly shaped objects.
\newblock \emph{IEEE Robotics and Automation Letters}, 8\penalty0
  (11):\penalty0 7026--7033, 2023.
\newblock \doi{10.1109/LRA.2023.3313877}.

\bibitem[Chen et~al.(2025)Chen, Barcel{\'o}, Saeed, Chen, Luces, and
  Hirata]{chen2025clothhand}
Xinhao Chen, Alberto El{\'i}as~Petrilli Barcel{\'o}, Shiry~Ghidary Saeed,
  Dayuan Chen, Jose Victorio~Salazar Luces, and Yasuhisa Hirata.
\newblock A versatile underactuated robotic hand for cloth manipulation.
\newblock \emph{IEEE Robotics and Automation Letters}, 10\penalty0
  (10):\penalty0 10314--10321, 2025.
\newblock \doi{10.1109/LRA.2025.3597486}.

\bibitem[He et~al.(2025)He, Yan, Jin, Fang, and Li]{he2025robotic}
Yiming He, Dengfeng Yan, Kaiyuan Jin, Bin Fang, and Shuguang Li.
\newblock A robotic hand with soft transmission systems for automated
  operations of computer mouse.
\newblock \emph{Advanced Intelligent Systems}, 7\penalty0 (8):\penalty0
  2400966, 2025.
\newblock \doi{10.1002/aisy.202400966}.

\bibitem[Weng(2025)]{weng2025bidexhand}
Zhengyang~Kris Weng.
\newblock Bidexhand: Design and evaluation of an open-source 16-dof biomimetic
  dexterous hand.
\newblock \emph{arXiv:2504.14712}, 2025.

\bibitem[Christoph et~al.(2025)Christoph, Eberlein, Katsimalis,
  et~al.]{christoph2025orca}
Clemens~C. Christoph, Maximilian Eberlein, Filippos Katsimalis, et~al.
\newblock Orca: An open-source, reliable, cost-effective, anthropomorphic
  robotic hand for uninterrupted dexterous task learning.
\newblock \emph{arXiv:2504.04259}, 2025.

\bibitem[Pan et~al.(2021)Pan, Garg, Anandkumar, and Zhu]{pan2021emergenthand}
Xinlei Pan, Animesh Garg, Animashree Anandkumar, and Yuke Zhu.
\newblock Emergent hand morphology and control from optimizing robust grasps of
  diverse objects.
\newblock In \emph{2021 IEEE International Conference on Robotics and
  Automation (ICRA)}, pages 7540--7547, 2021.
\newblock \doi{10.1109/ICRA48506.2021.9562092}.

\bibitem[Pinskier et~al.(2024)Pinskier, Wang, Liow, Xie, Kumar, Langelaar, and
  Howard]{pinskier2024diversity}
Josh Pinskier, Xing Wang, Lois Liow, Yue Xie, Prabhat Kumar, Matthijs
  Langelaar, and David Howard.
\newblock Diversity-based topology optimization of soft robotic grippers.
\newblock \emph{Advanced Intelligent Systems}, 6\penalty0 (4):\penalty0
  2300505, 2024.
\newblock \doi{10.1002/aisy.202300505}.

\bibitem[Qiu et~al.(2025)Qiu, Pa{\l}ucki, Ciebiera, Fija{\l}kowski, Cygan, and
  Kuci{\'n}ski]{qiu2024robomorph}
Kevin Qiu, W{\l}adys{\l}aw Pa{\l}ucki, Krzysztof Ciebiera, Pawe{\l}
  Fija{\l}kowski, Marek Cygan, and {\L}ukasz Kuci{\'n}ski.
\newblock Robomorph: Evolving robot morphology using large language models.
\newblock In \emph{ICLR 2025 Workshop on Foundation Models in the Wild
  (FM-Wild)}, 2025.

\bibitem[Chen et~al.(2026)Chen, Xu, Wang, Wang, and
  Huang]{chen2025naturalselection}
Changhe Chen, Xiaohao Xu, Xiangdong Wang, Jiaqi Wang, and Xiaonan Huang.
\newblock Natural selection via foundation models for soft robot evolution.
\newblock In \emph{2026 IEEE 9th International Conference on Soft Robotics
  (RoboSoft)}, pages 422--429, 2026.

\bibitem[Gao et~al.(2026)Gao, Li, Shi, Li, Zhang, Figueroa, and
  Jayaraman]{gao2025vlmgineer}
George~Jiayuan Gao, Tianyu Li, Junyao Shi, Yihan Li, Zizhe Zhang, Nadia
  Figueroa, and Dinesh Jayaraman.
\newblock Vlmgineer: Vision-language models as robotic toolsmiths.
\newblock In \emph{International Conference on Learning Representations}, 2026.

\bibitem[Qiu and Cygan(2026)]{qiu2025debate2create}
Kevin Qiu and Marek Cygan.
\newblock Debate2create: Robot co-design via multi-agent llm debate.
\newblock In \emph{Proceedings of the International Conference on Machine
  Learning (ICML)}, 2026.

\bibitem[Carlet et~al.(2025)Carlet, Gilday, and Hughes]{carlet2025behaviour}
Robin Carlet, Kieran Gilday, and Josie Hughes.
\newblock Behaviour range optimization of the dexterous robotic open parametric
  hand.
\newblock In \emph{2025 IEEE 8th International Conference on Soft Robotics},
  pages 1--7. IEEE, 2025.

\bibitem[Cutkosky(1989)]{cutkosky1989grasp}
Mark~R Cutkosky.
\newblock On grasp choice, grasp models, and the design of hands for
  manufacturing tasks.
\newblock \emph{IEEE Transactions on Robotics and Automation}, 5\penalty0
  (3):\penalty0 269--279, 1989.

\bibitem[Feix et~al.(2016)Feix, Romero, Schmiedmayer, Dollar, and
  Kragic]{feix2016grasp}
Thomas Feix, Javier Romero, Heinz-Bodo Schmiedmayer, Aaron~M Dollar, and Danica
  Kragic.
\newblock The grasp taxonomy of human grasp types.
\newblock \emph{IEEE Transactions on Human-Machine Systems}, 46\penalty0
  (1):\penalty0 66--77, 2016.

\bibitem[Kivell(2015)]{kivell2015evidence}
Tracy~L Kivell.
\newblock Evidence in hand: recent discoveries and the early evolution of human
  manual manipulation.
\newblock \emph{Philosophical Transactions of the Royal Society B: Biological
  Sciences}, 370\penalty0 (1682):\penalty0 20150105, 2015.

\bibitem[Jolliffe and Cadima(2016)]{jolliffe2016pca}
Ian~T. Jolliffe and Jorge Cadima.
\newblock Principal component analysis: a review and recent developments.
\newblock \emph{Philosophical Transactions of the Royal Society A:
  Mathematical, Physical and Engineering Sciences}, 374\penalty0
  (2065):\penalty0 20150202, 2016.
\newblock \doi{10.1098/rsta.2015.0202}.

\end{thebibliography}
\end{document}